\documentclass[11pt]{article}
\usepackage{acl2016}
\usepackage{times}
\usepackage{url}
\usepackage{latexsym}
\usepackage{color}
\usepackage{dcolumn}
\usepackage[pdftex]{graphicx}
\usepackage{arydshln}
\usepackage{caption,setspace}

\newcolumntype{d}{D{.}{.}{2.3}}           % alignment on decimal marker
\aclfinalcopy % Uncomment this line for the final submission

\title{End-to-End Relation Extraction using LSTMs \\ on Sequences and Tree Structures}

\author{Makoto Miwa\\
	    Toyota Technological Institute \\
	    Nagoya, 468-8511, Japan\\
	    {\tt makoto-miwa@toyota-ti.ac.jp}
	  \And
	Mohit Bansal\\
  	Toyota Technological Institute at Chicago \\ Chicago, IL, 60637, USA\\
  {\tt mbansal@ttic.edu}}

\date{}

\begin{document}

\maketitle

\begin{abstract}
We present a novel end-to-end neural model to extract entities and relations between them. Our recurrent neural network based model captures both word sequence and dependency tree substructure information by stacking bidirectional tree-structured LSTM-RNNs on bidirectional sequential LSTM-RNNs. This allows our model to jointly represent both entities and relations with shared parameters in a single model. We further encourage detection of entities during training and use of entity information in relation extraction via entity pretraining and scheduled sampling. Our model improves over the state-of-the-art feature-based model on end-to-end relation extraction, achieving 12.1\% and 5.7\% relative error reductions in F1-score on ACE2005 and ACE2004, respectively. 
We also show that our LSTM-RNN based model compares favorably to the state-of-the-art CNN based model (in F1-score) on nominal relation classification (SemEval-2010 Task 8). Finally, we present an extensive ablation analysis of several model components.
\end{abstract}

\section{Introduction}

Extracting semantic relations between entities in text is an important and well-studied task in information extraction and natural language processing (NLP). 
Traditional systems treat this task as a pipeline of two separated tasks, i.e., named entity recognition (NER)~\cite{nadeau2007survey,ratinov-roth:2009:CoNLL} and relation extraction~\cite{zelenko2003kernel,zhou-EtAl:2005:ACL}, but recent studies show that end-to-end (joint) modeling of entity and relation is important for high performance~\cite{li-ji:2014:P14-1,miwa-sasaki:2014:EMNLP2014} since relations interact closely with entity information. For instance, to learn that {\it Toefting} and {\it Bolton} have an {\it Organization-Affiliation (ORG-AFF)} relation in the sentence {\it Toefting transferred to Bolton}, the entity information that {\it Toefting} and {\it Bolton} are {\it Person} and {\it Organization} entities is important. Extraction of these entities is in turn encouraged by the presence of the context words {\it transferred to}, which indicate an employment relation.
Previous joint models have employed feature-based structured learning. An alternative approach to this end-to-end relation extraction task is to employ automatic feature learning via neural network (NN) based models. 

There are two ways to represent relations between entities using neural networks: 
recurrent/recursive neural networks (RNNs) and convolutional neural networks (CNNs). 
Among these, RNNs can {\it directly} represent essential linguistic structures, i.e., word sequences~\cite{hammerton2001clause} and constituent/dependency trees~\cite{tai-socher-manning:2015:ACL-IJCNLP}. Despite this representation ability, for relation classification tasks, the previously reported performance using long short-term memory (LSTM) based RNNs~\cite{xu-EtAl:2015:EMNLP2,li-EtAl:2015:EMNLP5} is worse than one using CNNs~\cite{dossantos-xiang-zhou:2015:ACL-IJCNLP}. These previous LSTM-based systems mostly include limited linguistic structures and neural architectures, and do not model entities and relations jointly. We are able to achieve improvements over state-of-the-art models via end-to-end modeling of entities and relations based on richer LSTM-RNN architectures that incorporate complementary linguistic structures. 

Word sequence and tree structure are known to be complementary information for extracting relations. 
For instance, dependencies between words are not enough to predict that {\it source} and {\it U.S.} have an {\it ORG-AFF} relation in the sentence {\it ``This is ...'', one U.S. source said}, and the context word {\it said} is required for this prediction. 
Many traditional, feature-based relation classification models extract features from both sequences and parse trees~\cite{zhou-EtAl:2005:ACL}. 
However, previous RNN-based models focus on only one of these linguistic structures~\cite{socher-EtAl:2012:EMNLP-CoNLL}. 

We present a novel end-to-end model to extract relations between entities on both word sequence and dependency tree structures. Our model allows joint modeling of entities and relations in a single model by using both bidirectional sequential (left-to-right and right-to-left) and bidirectional tree-structured (bottom-up and top-down) LSTM-RNNs. 
Our model first detects entities and then extracts relations between the detected entities using a single incrementally-decoded NN structure, and the NN parameters are jointly updated using both entity and relation labels. 
Unlike traditional incremental end-to-end relation extraction models, our model further incorporates two enhancements into training: entity pretraining, which pretrains the entity model, and scheduled sampling~\cite{bengio2015scheduled}, which replaces (unreliable) predicted labels with gold labels in a certain probability. These enhancements alleviate the problem of low-performance entity detection in early stages of training, as well as allow entity information to further help downstream relation classification.

On end-to-end relation extraction, we improve over the state-of-the-art feature-based model, with 12.1\% (ACE2005) and 5.7\% (ACE2004) relative error reductions in F1-score. 
On nominal relation classification (SemEval-2010 Task 8), our model compares favorably to the state-of-the-art CNN-based model in F1-score. 
Finally, we also ablate and compare our various model components, which leads to some key findings (both positive and negative) about the contribution and effectiveness of different RNN structures, input dependency relation structures, different parsing models, external resources, and joint learning settings.

\section{Related Work}

LSTM-RNNs have been widely used for sequential labeling, such as clause identification~\cite{hammerton2001clause}, phonetic labeling~\cite{graves2005framewise}, 
and NER~\cite{Hammerton:2003:CONLL}.
Recently, \newcite{huang2015bidirectional} showed that building a conditional
random field (CRF) layer on top of bidirectional LSTM-RNNs performs comparably to the state-of-the-art methods in the part-of-speech (POS) tagging, chunking, and NER.

For relation classification, in addition to traditional feature/kernel-based approaches~\cite{zelenko2003kernel,bunescu2005shortest}, several neural models have been proposed in the SemEval-2010 Task 8~\cite{hendrickx-EtAl:2010:SemEval}, including embedding-based models~\cite{hashimoto-EtAl:2015:CoNLL}, CNN-based models~\cite{dossantos-xiang-zhou:2015:ACL-IJCNLP}, and RNN-based models~\cite{socher-EtAl:2012:EMNLP-CoNLL}.
Recently, \newcite{xu-EtAl:2015:EMNLP1} and \newcite{xu-EtAl:2015:EMNLP2} showed that the shortest dependency paths between relation arguments, which were used in feature/kernel-based systems~\cite{bunescu2005shortest}, are also useful in NN-based models.
\newcite{xu-EtAl:2015:EMNLP2} also showed that LSTM-RNNs are useful for relation
classification, but the performance was worse than CNN-based models.
\newcite{li-EtAl:2015:EMNLP5} compared separate sequence-based and tree-structured LSTM-RNNs on relation classification, using basic RNN model structures. 

Research on tree-structured LSTM-RNNs~\cite{tai-socher-manning:2015:ACL-IJCNLP} fixes the direction of information propagation from bottom to top, and also cannot handle an arbitrary number of typed children as in a typed dependency tree. Furthermore, no RNN-based relation classification model simultaneously uses word sequence and dependency tree information. 
We propose several such novel model structures and training settings, investigating the simultaneous use of bidirectional sequential and bidirectional tree-structured LSTM-RNNs to jointly capture linear and dependency context for end-to-end extraction of relations between entities. 

As for end-to-end (joint) extraction of relations between entities, all existing models are feature-based systems (and no NN-based model has been proposed). Such models include structured
prediction~\cite{li-ji:2014:P14-1,miwa-sasaki:2014:EMNLP2014}, integer linear
programming~\cite{RothYi07,yang-cardie:2013:ACL2013}, card-pyramid
parsing~\cite{kate-mooney:2010:CONLL}, and global probabilistic graphical
models~\cite{yu-lam:2010:POSTERS1,singh2013joint}. Among these, structured prediction methods are state-of-the-art on several corpora.
We present an improved, NN-based alternative for the end-to-end relation extraction. 

\begin{figure*}[t!]
\centering
\includegraphics[width=.8\linewidth]{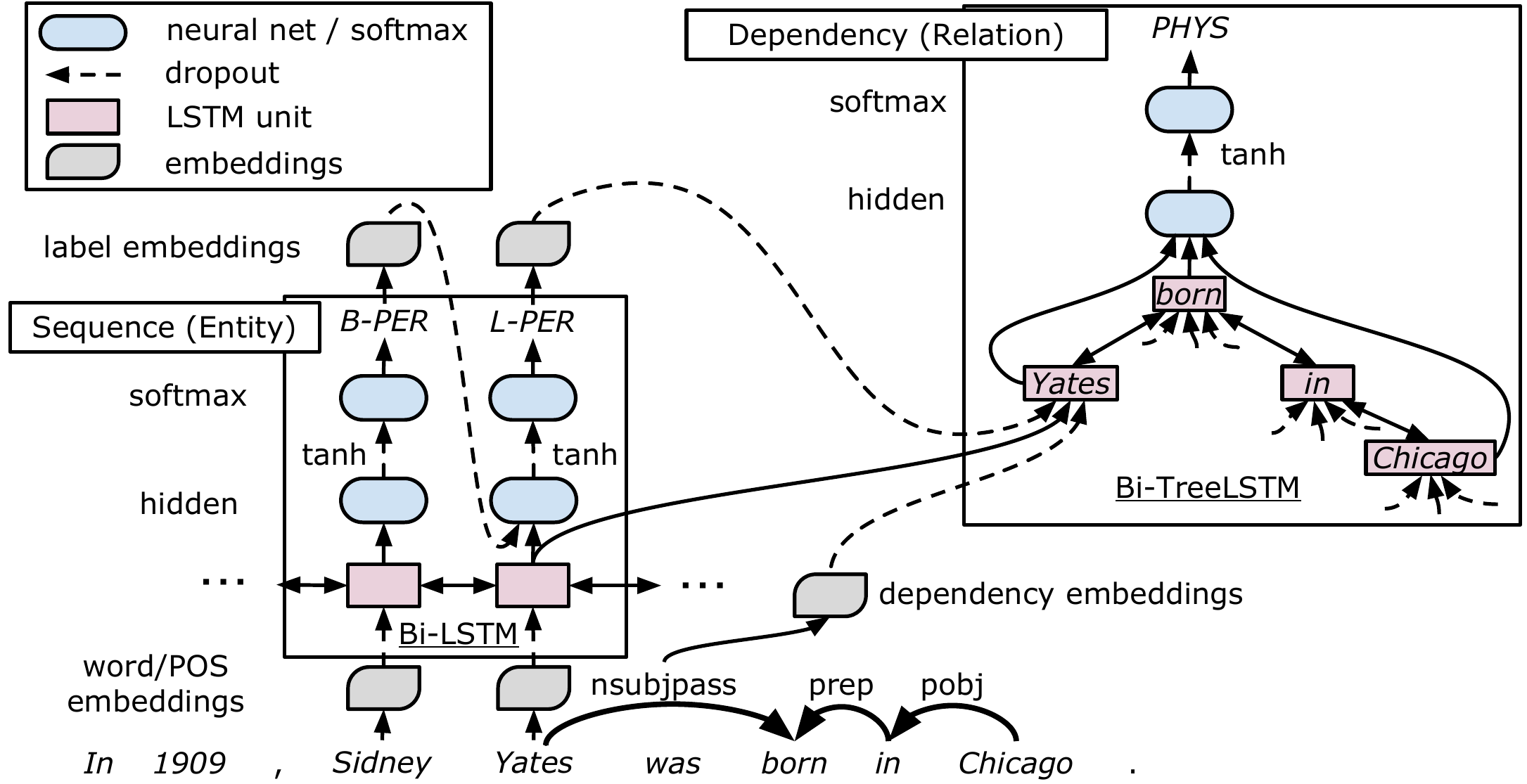}
\caption{Our incrementally-decoded end-to-end relation extraction model, with bidirectional sequential and bidirectional tree-structured LSTM-RNNs.}
\label{fig:overview}
\end{figure*}

\section{Model}

We design our model with LSTM-RNNs that represent both word sequences and dependency tree structures, and perform end-to-end extraction of relations between entities on top of these RNNs. 
Fig.~\ref{fig:overview} illustrates the overview of the model.
The model mainly consists of three representation layers: 
a word embeddings layer (embedding layer), a word sequence based LSTM-RNN layer (sequence layer), and finally a dependency subtree based LSTM-RNN layer (dependency layer). 
During decoding, we build greedy, left-to-right entity detection on the sequence layer and realize relation classification on the dependency layers, where each subtree based LSTM-RNN corresponds to a relation candidate between two detected entities. 
After decoding the entire model structure, we update the parameters simultaneously via back-propagation through time (BPTT)~\cite{werbos1990backpropagation}. 
The dependency layers are stacked on the sequence layer, so the embedding and sequence layers are {\bf shared} by both entity detection and relation classification, and the shared parameters are affected by both entity and relation labels. 

\subsection{Embedding Layer}

The embedding layer handles embedding representations. $n_w$, $n_p$, $n_d$ and $n_e$-dimensional vectors $v^{(w)}$, $v^{(p)}$, $v^{(d)}$ and $v^{(e)}$ are
embedded to words, part-of-speech (POS) tags, dependency types, and entity labels, respectively. 

\subsection{Sequence Layer}

The sequence layer represents words in a linear sequence using the representations from the embedding layer. This layer represents sentential context information and maintains entities, as shown in bottom-left part of Fig.~\ref{fig:overview}.

We represent the word sequence in a sentence with bidirectional LSTM-RNNs~\cite{graves2013speech}.
The LSTM unit at $t$-th word consists of a collection of $n_{l_s}$-dimensional
vectors: an input gate $i_t$, a forget gate $f_t$, an output gate $o_t$, a
memory cell $c_t$, and a hidden state $h_t$.
The unit receives an $n$-dimensional input vector $x_t$, the previous hidden state $h_{t-1}$, and the memory cell $c_{t-1}$, and calculates the new vectors using the following equations:
\vspace{-0.15cm}\begin{eqnarray}
i_t&=&\sigma\left(W^{(i)}x_t+U^{(i)}h_{t-1} + b^{(i)}\right),\\
f_t&=&\sigma\left(W^{(f)}x_t+U^{(f)}h_{t-1} + b^{(f)}\right),\nonumber\\
o_t&=&\sigma\left(W^{(o)}x_t+U^{(o)}h_{t-1} + b^{(o)}\right),\nonumber\\
u_t&=&\tanh\left(W^{(u)}x_t+U^{(u)}h_{t-1} + b^{(u)}\right),\nonumber\\
c_t&=&i_t {\odot} u_t + f_t {\odot} c_{t-1},\nonumber\\
h_t&=&o_t {\odot} \tanh(c_t),\nonumber
\end{eqnarray}
where $\sigma$ denotes the logistic function, $\odot$ denotes element-wise multiplication, $W$ and $U$ are weight matrices, and $b$ are bias vectors.
The LSTM unit at $t$-th word receives the concatenation of word and POS embeddings as its input vector: $x_t=\left[v^{(w)}_{t}; v^{(p)}_{t}\right]$.
We also concatenate the hidden state vectors of the two directions' LSTM units corresponding to each word (denoted as $\overrightarrow{h_t}$ and $\overleftarrow{h_t}$) as its output vector, $s_t = \left[\overrightarrow{h_t}; \overleftarrow{h_t}\right]$, and pass it to the subsequent layers.

\subsection{Entity Detection}

We treat entity detection as a sequence labeling task. We assign an entity tag to each word using a commonly used encoding scheme BILOU (Begin, Inside, Last, Outside, Unit)~\cite{ratinov-roth:2009:CoNLL}, where each entity tag represents the entity type and the position of a word in the entity. For example, in Fig.~\ref{fig:overview}, we assign {\it B-PER} and {\it L-PER} (which denote the beginning and last words of a person entity type, respectively) to each word in {\it Sidney Yates} to represent this phrase as a {\it PER} (person) entity type.

We perform entity detection on top of the sequence layer. We employ a two-layered NN with an $n_{h_e}$-dimensional hidden layer $h^{(e)}$ and a softmax output layer for entity detection.
\begin{eqnarray}
h^{(e)}_t &=& \tanh \left( W^{(e_h)} [s_t; v^{(e)}_{t-1}] + b^{(e_h)} \right)\\
y_t &=& \mbox{softmax}\left( W^{(e_y)}h^{(e)}_t + b^{(e_y)} \right)
\end{eqnarray}
Here, $W$ are weight matrices and $b$ are bias vectors. 

We assign entity labels to words in a greedy, left-to-right manner.\footnote{We also tried beam search but this did not show improvements in initial experiments.}
During this decoding, we use the predicted label of a word to predict the label of the next word so as to take label dependencies into account. The NN above receives the concatenation of its corresponding outputs in the sequence layer and the label embedding for its previous word (Fig.~\ref{fig:overview}). 

\subsection{Dependency Layer}

The dependency layer represents a relation between a pair of two target words (corresponding to a relation candidate in relation classification) in the dependency tree, and is in charge of relation-specific representations, as is shown in top-right part of Fig.~\ref{fig:overview}. This layer mainly focuses on the \emph{shortest path} between a pair of target words in the dependency tree (i.e., the path between the least common node and the two target words) since these paths are shown to be effective in relation classification~\cite{xu-EtAl:2015:EMNLP1}. For example, we show the shortest path between {\it Yates} and {\it Chicago} in the bottom of Fig.~\ref{fig:overview}, and this path well captures the key phrase of their relation, i.e., {\it born in}.

We employ bidirectional tree-structured LSTM-RNNs (i.e., bottom-up and top-down) to represent a relation candidate by capturing the dependency structure around the target
word pair. This bidirectional structure propagates to each node not only the information from the leaves but also information from the root. This is especially important for relation classification, which makes use of argument nodes near the bottom of the tree, and our top-down LSTM-RNN sends information from the top of the tree to such near-leaf nodes (unlike in standard bottom-up LSTM-RNNs).\footnote{We also tried to use one LSTM-RNN by connecting the root~\cite{NIPS2014_5275}, but preparing two LSTM-RNNs showed slightly better performance in our initial experiments.}
Note that the two variants of tree-structured LSTM-RNNs by \newcite{tai-socher-manning:2015:ACL-IJCNLP} are not able to represent our target structures which have a variable number of typed children: the Child-Sum Tree-LSTM does not deal with types and the $N$-ary Tree assumes a fixed number of children. 
We thus propose a new variant of tree-structured LSTM-RNN that shares weight matrices $U$s for same-type children and also allows variable number of children. For
this variant, we calculate $n_{l_t}$-dimensional vectors in the LSTM unit at $t$-th node with $C(t)$ children using following equations:
\vspace{-0.15cm}\begin{eqnarray}
&&\hspace{-20pt}i_t=\sigma\left(W^{(i)}x_t+\sum_{l{\in}C(t)}U_{m(l)}^{(i)}h_{tl} +
b^{(i)}\right),\\
&&\hspace{-20pt}f_{tk}=\sigma\left(W^{(f)}x_t+\sum_{l{\in}C(t)}U_{m(k)m(l)}^{(f)}h_{tl} +
b^{(f)}\right),\hspace{-80pt}\nonumber\\
&&\hspace{-20pt}o_t=\sigma\left(W^{(o)}x_t+\sum_{l{\in}C(t)}U_{m(l)}^{(o)}h_{tl} +
b^{(o)}\right),\nonumber\\
&&\hspace{-20pt}u_t=\tanh\left(W^{(u)}x_t+\sum_{l{\in}C(t)}U_{m(l)}^{(u)}h_{tl} +
b^{(u)}\right),\nonumber\\
&&\hspace{-20pt}c_t=i_t {\odot} u_t + \sum_{l{\in}C(t)} f_{tl} {\odot} c_{tl},\nonumber\\
&&\hspace{-20pt}h_t=o_t {\odot} \tanh(c_t),\nonumber
\end{eqnarray}
where $m(\cdot)$ is a type mapping function. 

To investigate appropriate structures to represent relations between two target word pairs, we experiment with three structure options. We primarily employ the shortest path structure ({\bf SPTree}), which captures the core dependency path between a target word pair and is widely used in relation classification models, e.g., ~\cite{bunescu2005shortest,xu-EtAl:2015:EMNLP1}.
We also try two other dependency structures: {\bf SubTree} and {\bf FullTree}. 
SubTree is the subtree under the lowest common ancestor of the target word pair. This provides additional modifier information to the path and the word pair in SPTree. FullTree is the full dependency tree. This captures context from the entire sentence. While we use one node type for SPTree, we define two node types for SubTree and FullTree, i.e., one for nodes on shortest paths and one for all other nodes. We use the type mapping function $m(\cdot)$ to distinguish these two nodes types.

\subsection{Stacking Sequence and Dependency Layers} 

We stack the dependency layers (corresponding to relation candidates) on top of the sequence layer to incorporate both word sequence and dependency tree structure information into the output.
The dependency-layer LSTM unit at the $t$-th word receives as input $
x_t=\left[s_t; v^{(d)}_{t}; v^{(e)}_{t} \right]$, i.e., the concatenation of its corresponding hidden state
vectors $s_t$ in the sequence layer, dependency type embedding $v^{(d)}_t$ (denotes the type of dependency to the parent{\footnote{We use the dependency to the parent since the number of children varies. Dependency types can also be incorporated into $m(\cdot)$, but this did not help in initial experiments.}}), and label embedding $v^{(e)}_t$ (corresponds to the predicted entity label).

\subsection{Relation Classification}

We incrementally build relation candidates using all possible combinations of the last words of detected entities, i.e., words with L or U labels in the BILOU scheme, during decoding. For instance, in Fig.~\ref{fig:overview}, we build a relation candidate using {\it Yates} with an {\it L-PER} label and {\it Chicago} with an {\it U-LOC} label.
For each relation candidate, we realize the dependency layer $d_p$ (described above) corresponding to the path between the word pair $p$ in the relation candidate, and the NN receives a relation candidate vector constructed from the output of the dependency tree layer, and predicts its relation label. We treat a pair as a negative relation when the detected entities are wrong or when the pair has no relation. We represent relation labels by type and direction, except for negative relations that have no direction. 

The relation candidate vector is constructed as the concatenation $d_p = \left[{\uparrow}h_{p_A}; {\downarrow}h_{p_1}; {\downarrow}h_{p_2}\right]$, where ${\uparrow}h_{p_A}$ is the hidden state vector of the top LSTM unit in the bottom-up LSTM-RNN (representing the lowest common ancestor of the target word pair $p$), and ${\downarrow}h_{p_1}$, ${\downarrow}h_{p_2}$ are the hidden state vectors of the two LSTM units representing the first and second target words in the top-down LSTM-RNN.\footnote{Note that the order of the target words corresponds to the direction of the relation, not the positions in the sentence.} All the corresponding arrows are shown in Fig.~\ref{fig:overview}.

Similarly to the entity detection, we employ a two-layered NN with an $n_{h_r}$-dimensional hidden layer $h^{(r)}$ and a softmax output layer (with weight matrices $W$, bias vectors $b$).
\begin{eqnarray}
h^{(r)}_p &=& \tanh \left( W^{(r_h)} d_p + b^{(r_h)} \right)\\
y_p &=& \mbox{softmax}\left( W^{(r_y)}h^{(r)}_t + b^{(r_y)} \right)
\end{eqnarray}

We construct the input $d_p$ for relation classification from tree-structured LSTM-RNNs stacked on sequential LSTM-RNNs, 
so the contribution of sequence layer to the input is indirect. 
Furthermore, our model uses words for representing entities, so it cannot fully use the entity information.  
To alleviate these problems, we directly concatenate the average of hidden state vectors for each entity from the sequence layer to the input $d_p$ to relation classification, i.e., $d_p'=\left[d_p;\frac{1}{|I_{p_1}|}\sum_{i \in I_{p_1}} s_{i};\frac{1}{|I_{p_2}|}\sum_{i \in I_{p_2}} s_{i}\right]$ ({\bf Pair}), where $I_{p_1}$ and $I_{p_2}$ represent sets of word indices in the first and second entities.\footnote{Note that we do not show this {\bf Pair} in Fig.\ref{fig:overview} for simplicity.} 

Also, we assign two labels to each word pair in prediction since we consider both left-to-right and right-to-left directions. When the predicted labels are inconsistent, we select the positive and more confident label, similar to \newcite{xu-EtAl:2015:EMNLP1}.

\subsection{Training}

We update the model parameters including weights, biases, and embeddings by BPTT and Adam~\cite{kingma2014adam} with
gradient clipping, parameter averaging, and L2-regularization (we regularize weights $W$ and $U$, not the bias terms $b$). We also apply dropout~\cite{srivastava2014dropout} 
to the embedding layer and to the final hidden layers for entity detection and relation classification.

We employ two enhancements, \textbf{scheduled sampling}~\cite{bengio2015scheduled} and \textbf{entity pretraining}, to alleviate the problem of unreliable prediction of entities in the early stage of training, and to encourage building positive relation instances from the detected entities. In scheduled sampling, we use gold labels as prediction in the probability of $\epsilon_i$ that depends on the number of epochs $i$ during training if the gold labels are legal.
As for $\epsilon_i$, we choose the inverse sigmoid decay $\epsilon_i=k/(k+\exp(i/k))$, where $k (\geq 1)$ is a hyper-parameter that adjusts how often we use the gold labels as prediction. Entity pretraining is inspired by ~\cite{PentinaSL15}, and we pretrain the entity detection model using the training data before training the entire model parameters. 

\section{Results and Discussion}

\subsection{Data and Task Settings}

We evaluate on three datasets:  
ACE05 and ACE04 for end-to-end relation extraction, and SemEval-2010 Task 8 for relation classification. We use the first two datasets as our primary target, and use the last one to thoroughly analyze and ablate the relation classification part of our model. 

{\bf ACE05} defines 7 coarse-grained entity types and 6 coarse-grained relation types between entities. We use the same data splits, preprocessing, and task settings as \newcite{li-ji:2014:P14-1}. 
We report the primary micro F1-scores as well as micro precision and recall on both entity and relation extraction to better explain model performance. 
We treat an entity as correct when its type and the region of its
head are correct. We treat a relation as correct when its type and argument entities are correct; we thus treat all non-negative relations on wrong entities as false positives.

{\bf ACE04} defines the same 7 coarse-grained entity types as ACE05~\cite{doddington:5:2004:lrec2004}, but defines 7 coarse-grained relation types.
We follow the cross-validation setting of \newcite{chan-roth:2011:ACL-HLT2011} and \newcite{li-ji:2014:P14-1}, and the preprocessing and evaluation metrics of ACE05.

{\bf SemEval-2010 Task 8} defines 9 relation types between nominals and a tenth type {\it Other} when two nouns have none of these relations~\cite{hendrickx-EtAl:2010:SemEval}. We treat this {\it Other} type as a negative relation type, and no direction is considered.
The dataset consists of 8,000 training and 2,717 test sentences, and each sentence is annotated with a relation between two given nominals.
We randomly selected 800 sentences from the training set as our development set. We followed the official task setting, and report the official
macro-averaged F1-score (Macro-F1) on the 9 relation types. 

For more details of the data and task settings, please refer to the supplementary material.

\begin{table*}[t!]
\centering
\begin{tabular}{|llrrrrrr|}
\hline
Corpus & Settings & \multicolumn{3}{c}{Entity} & \multicolumn{3}{c|}{Relation} \\
       &           & P  & R & F1 & P & R & F1\\
\hline
\hline
ACE05 & Our Model (SPTree) & 0.829 & \bf{0.839} & \bf{0.834} & 0.572 & \bf{0.540} & \bf{0.556} \\
& \newcite{li-ji:2014:P14-1} & \bf{0.852} & 0.769 & 0.808 & \bf{0.654} & 0.398 & 0.495\\
\hline
ACE04 & Our Model (SPTree) & 0.808 & \bf{0.829} & \bf{0.818} & 0.487 & \bf{0.481} & \bf{0.484} \\
& \newcite{li-ji:2014:P14-1} & \bf{0.835} & 0.762 & 0.797 & \bf{0.608} & 0.361 & 0.453 \\
\hline
\end{tabular}
\caption{Comparison with the state-of-the-art on the ACE05 test set and ACE04 dataset.}
\label{tbl:ace04-05-test}
\end{table*}
\begin{table*}[t!]
\centering
\begin{tabular}{|lrrrrrl|}
\hline
Settings & \multicolumn{3}{c}{Entity} & \multicolumn{3}{c|}{Relation} \\
             & P  & R & F1 & P & R & \multicolumn{1}{c|}{F1}\\
\hline
\hline
Our Model (SPTree) & 0.815 & 0.821 & 0.818 & 0.506 & 0.529 & 0.518 \\
$-$Entity pretraining (EP) & 0.793 & 0.798 & 0.796 & 0.494 & 0.491 & 0.492* \\
$-$Scheduled sampling (SS) & 0.812 & 0.818 & 0.815 & 0.522 & 0.490 & 0.505 \\
$-$Label embeddings (LE) & 0.811 & 0.821 & 0.816 & 0.512 & 0.499 & 0.505 \\
$-$Shared parameters (Shared) & 0.796 & 0.820 & 0.808 & 0.541 & 0.482 & 0.510 \\
$-$EP, SS & 0.781 & 0.804 & 0.792 & 0.509 & 0.479 & 0.494* \\
%$-$EP, SS, LE & 0.795 & 0.809 & 0.802 & 0.512 & 0.465 & 0.487 \\
$-$EP, SS, LE, Shared & 0.800 & 0.815 & 0.807 & 0.520 & 0.452 & 0.484** \\
\hline
\end{tabular}
\caption{Ablation tests on the ACE05 development dataset. * denotes significance at p$<$0.05, ** denotes p$<$0.01.}
\label{tbl:ace05-dev}
\end{table*}
\begin{table*}[t!]
\centering
\begin{tabular}{|lrrrrrl|}
\hline
Settings & \multicolumn{3}{c}{Entity} & \multicolumn{3}{c|}{Relation} \\
             & P  & R & F1 & P & R & \multicolumn{1}{c|}{F1}\\
\hline
\hline
SPTree & 0.815 & 0.821 & 0.818 & 0.506 & 0.529 & 0.518 \\
SubTree & 0.812 & 0.818 & 0.815 & 0.525 & 0.506 & 0.515 \\
FullTree & 0.806 & 0.816 & 0.811 & 0.536 & 0.507 & 0.521 \\ 
SubTree (-SP) & 0.803 & 0.816 & 0.810 & 0.533 & 0.495 & 0.514 \\
FullTree (-SP) & 0.804 & 0.817 & 0.811 & 0.517 & 0.470 & 0.492*\\
\hline
Child-Sum & 0.806 & 0.819 & 0.8122 & 0.514 & 0.499 & 0.506 \\
SPSeq & 0.801 & 0.813 & 0.807 & 0.500 & 0.523 & 0.511 \\
SPXu & 0.809 & 0.818 & 0.813 & 0.494 & 0.522 & 0.508 \\
\hline
\end{tabular}
\caption{Comparison of LSTM-RNN structures on the ACE05 development dataset. }
\label{tbl:ace05-lstm-dev}
\end{table*}

\subsection{Experimental Settings}

We implemented our model using the {\it cnn}
library.\footnote{{\scriptsize\url{https://github.com/clab/cnn}}}
We parsed the texts using the Stanford neural dependency
parser\footnote{{\scriptsize\url{http://nlp.stanford.edu/software/stanford-corenlp-full-2015-04-20.zip}}}~\cite{chen-manning:2014:EMNLP2014} with the original Stanford Dependencies.
Based on preliminary tuning, we fixed embedding dimensions $n_w$ to 200,
$n_p$, $n_d$, $n_e$ to 25, and
dimensions of intermediate layers ($n_{l_s}$, $n_{l_t}$ of LSTM-RNNs and $n_{h_e}$, $n_{h_r}$ of hidden layers) to 100.
We initialized word vectors via word2vec~\cite{mikolov2013distributed} trained on Wikipedia\footnote{{\scriptsize\url{https://dumps.wikimedia.org/enwiki/20150901/}}} and randomly initialized all other parameters.
We tuned hyper-parameters using development sets for ACE05 and SemEval-2010 Task 8 to achieve high primary (Micro- and Macro-) F1-scores.\footnote{We did not tune the precision-recall trade-offs, but doing so can specifically improve precision further.} For ACE04, we directly employed the best parameters for ACE05. The hyper-parameter settings are shown in the supplementary material. 
For SemEval-2010 Task 8, we also omitted the entity detection and label embeddings since only target nominals are annotated and the task defines no entity types. 
Our statistical significance results are based on the Approximate Randomization (AR) test~\cite{noreen89:_comput_inten_method_for_testin_hypot}.

\subsection{End-to-end Relation Extraction Results}

Table~\ref{tbl:ace04-05-test} compares our model with the state-of-the-art feature-based model of \newcite{li-ji:2014:P14-1}\footnote{Other work on ACE is not comparable or performs worse than the model by \newcite{li-ji:2014:P14-1}.} on final test sets, and shows that our model performs better than the state-of-the-art model. 

To analyze the contributions and effects of the various components of our end-to-end relation extraction model, 
we perform ablation tests on the ACE05 development set (Table~\ref{tbl:ace05-dev}).
The performance slightly degraded without scheduled sampling, and the performance significantly degraded when we removed entity pretraining or removed both (p$<$0.05).
This is reasonable because the model can only create relation instances when both of the entities are found and, without these enhancements, it may get too late to find some relations. 
Removing label embeddings did not affect the entity detection performance, but this degraded the recall in relation classification. This indicates that entity label information is helpful in detecting relations. 

We also show the performance without sharing parameters, i.e., embedding and sequence layers, for detecting entities and relations ({\bf $-$Shared parameters}); we first train the entity detection model, detect entities with the model, and build a {\it separate} relation extraction model using the detected entities, i.e., without entity detection. This setting can be regarded as a pipeline model since two separate models are trained sequentially. 
Without the shared parameters, both the performance in entity detection and relation classification
drops slightly, although the differences are not significant.
When we removed all the enhancements, i.e., scheduled sampling, entity pretraining, label embedding, and shared parameters, the performance is significantly worse than SPTree (p$<$0.01), showing that these enhancements provide complementary benefits to end-to-end relation extraction.

Next, we show the performance with different LSTM-RNN structures in Table~\ref{tbl:ace05-lstm-dev}.
We first compare the three input dependency structures (SPTree, SubTree, FullTree) 
for tree-structured LSTM-RNNs. 
Performances on these three structures are almost same when we distinguish the nodes in the shortest paths from other nodes, but when we do not distinguish them (-SP), the information outside of the shortest path, i.e., FullTree (-SP), significantly hurts performance (p$<$0.05).
We then compare our tree-structured LSTM-RNN (SPTree) with the Child-Sum tree-structured LSTM-RNN on the shortest path of~\newcite{tai-socher-manning:2015:ACL-IJCNLP}. Child-Sum performs worse than our SPTree model, but not with as big of a decrease as above.
This may be because the difference in the models appears only on nodes that have multiple children and all the nodes except for the least common node have one child.

We finally show results with two counterparts of sequence-based LSTM-RNNs using the shortest path (last two rows in Table~\ref{tbl:ace05-lstm-dev}).
{\bf SPSeq} is a bidirectional LSTM-RNN on the shortest path. The LSTM unit
receives input from the sequence layer concatenated with embeddings for the surrounding dependency types and directions. We concatenate the outputs of the two RNNs for the relation candidate. 
{\bf SPXu} is our adaptation of the shortest path LSTM-RNN proposed by \newcite{xu-EtAl:2015:EMNLP2} to match our sequence-layer based model.\footnote{This is different from the original one in that we use the sequence layer and we concatenate the embeddings for the input, while the original one prepared individual LSTM-RNNs for different inputs and concatenated their outputs.}
This has two LSTM-RNNs for the left and right subpaths of the shortest path. We first calculate the max pooling of the LSTM units for each of these two RNNs, and then concatenate the outputs of the pooling for the relation candidate. 
The comparison with these sequence-based LSTM-RNNs indicates that a tree-structured LSTM-RNN is comparable to sequence-based ones in representing shortest paths. 

Overall, the performance comparison of the LSTM-RNN structures in Table~\ref{tbl:ace05-lstm-dev} show that for end-to-end relation extraction, \emph{selecting the appropriate tree structure representation of the input (i.e., the shortest path) is more important than the choice of the LSTM-RNN structure on that input (i.e., sequential versus tree-based).}

\subsection{Relation Classification Analysis Results}

To thoroughly analyze the relation classification part alone, e.g., comparing different LSTM structures, architecture components such as hidden layers and input information, and classification task settings, we use the SemEval-2010 Task 8. This dataset, often used to evaluate NN models for relation classification, annotates only relation-related nominals (unlike ACE datasets), so we can focus cleanly on the relation classification part. 

We first report official test set results in Table~\ref{tbl:official_semeval}.
Our novel LSTM-RNN model is comparable to both the state-of-the-art CNN-based models on this task with or without external sources, i.e., WordNet, unlike the previous best LSTM-RNN model~\cite{xu-EtAl:2015:EMNLP2}.\footnote{When incorporating WordNet information into our model, we prepared embeddings for WordNet hypernyms extracted by SuperSenseTagger~\cite{ciaramita-altun:2006:EMNLP} and concatenated the embeddings to the input vector (the concatenation of word and POS embeddings) of the sequence LSTM. We tuned the dimension of the WordNet embeddings and set it to 15 using the development dataset.}

\begin{table}[t!]
\centering
\begin{tabular}{|lr|}
\hline
Settings & Macro-F1 \\
\hline
\hline
\multicolumn{2}{|c|}{No External Knowledge Resources}\\
\hline
Our Model (SPTree) &  \bf{0.844} \\
\newcite{dossantos-xiang-zhou:2015:ACL-IJCNLP} & 0.841 \\
\newcite{xu-EtAl:2015:EMNLP1} & 0.840 \\
\hline
\hline
\multicolumn{2}{|c|}{+WordNet}\\
\hline
Our Model (SPTree + WordNet) & 0.855 \\
\newcite{xu-EtAl:2015:EMNLP1} & \bf{0.856} \\
\newcite{xu-EtAl:2015:EMNLP2} & 0.837 \\
\hline
\end{tabular}
\caption{Comparison with state-of-the-art models 
on SemEval-2010 Task 8 test-set.}
\label{tbl:official_semeval}
\end{table}

\begin{table}[t!]
\centering
\begin{tabular}{|ld|}
\hline
Settings & \multicolumn{1}{r|}{Macro-F1} \\
\hline
\hline
SPTree & 0.851 \\
SubTree & 0.839 \\
FullTree & 0.829* \\
SubTree (-SP) & 0.840 \\
FullTree (-SP) & 0.828* \\
\hline
Child-Sum & 0.838 \\
SPSeq & 0.844 \\
SPXu & 0.847 \\
\hline
\end{tabular}
\caption{Comparison of LSTM-RNN structures on SemEval-2010 Task 8 development set.}
\label{tbl:semeval-relmodel-dev}
\end{table}

Next, we compare different LSTM-RNN structures in Table~\ref{tbl:semeval-relmodel-dev}.
As for the three input dependency structures (SPTree, SubTree, FullTree), FullTree performs significantly worse than other structures regardless of whether or not we distinguish the nodes in the shortest paths from the other nodes, which hints that the information outside of the shortest path significantly hurts the performance (p$<$0.05). 
We also compare our tree-structured LSTM-RNN (SPTree) with sequence-based LSTM-RNNs (SPSeq and SPXu) and tree-structured LSTM-RNNs (Child-Sum). All these LSTM-RNNs perform slightly worse than our SPTree model, but the differences are small.

Overall, for relation classification, although the performance comparison of the LSTM-RNN structures in Table~\ref{tbl:semeval-relmodel-dev} produces different results on FullTree as compared to the results on ACE05 in Table~\ref{tbl:ace05-lstm-dev}, the trend still holds that selecting the appropriate tree structure representation of the input is more important than the choice of the LSTM-RNN structure on that input.

\begin{table}[t!]
\centering
\begin{tabular}{|ld|}
\hline
Settings & \multicolumn{1}{r|}{Macro-F1} \\
\hline
\hline
SPTree & 0.851 \\
\hline
$-$Hidden layer & 0.839 \\
\hline
$-$Sequence layer & 0.840 \\
$-$Pair & 0.844 \\
$-$Pair, Sequence layer & 0.827* \\
%$-$Pair, Sequence \& Hidden layers & 0.831* \\
\hline
Stanford PCFG & 0.844 \\
\hline
$+$WordNet & 0.854 \\
\hline
Left-to-right candidates & 0.843 \\
Neg. sampling~\cite{xu-EtAl:2015:EMNLP1} & 0.848 \\
\hline
\end{tabular}
\caption{Model setting ablations on SemEval-2010 development set.}
\label{tbl:semeval-rel-options}
\end{table}

Finally, Table~\ref{tbl:semeval-rel-options} summarizes the contribution of several model components and training settings on SemEval relation classification. 
We first remove the hidden layer by directly connecting the LSTM-RNN layers to the softmax layers, and found that this slightly degraded performance, but the difference was small.
We then skip the sequence layer and directly use the word and POS embeddings for the dependency layer. Removing the sequence layer\footnote{Note that this setting still uses some sequence layer information since it uses the entity-related information (Pair).} or entity-related information from the sequence layer ($-$Pair) slightly degraded performance, and, on removing both, 
the performance dropped significantly (p$<$0.05). 
This indicates that the sequence layer is necessary 
but the last words of nominals are almost enough for 
expressing the relations in this task.

When we replace the Stanford neural dependency parser with the Stanford lexicalized PCFG parser (Stanford PCFG), the performance slightly dropped, but the difference was small. 
This indicates that the selection of parsing models is not critical.
We also included WordNet, and this slightly improved the performance ($+$WordNet), but the difference was small.
Lastly, for the generation of relation candidates, generating only left-to-right
candidates slightly degraded the performance, 
but the difference was small and hence the creation of right-to-left candidates was not critical. 
Treating the inverse relation candidate as a negative instance (Negative sampling) also performed comparably to other generation methods in our model (unlike \newcite{xu-EtAl:2015:EMNLP1}, which showed a significance improvement over generating only left-to-right candidates). 

\section{Conclusion}

We presented a novel end-to-end relation extraction model that represents both word sequence and dependency tree structures by using bidirectional sequential and bidirectional tree-structured LSTM-RNNs. This allowed us to represent both entities and relations in a single model, achieving gains over the state-of-the-art, feature-based system on end-to-end relation extraction (ACE04 and ACE05), and showing favorably comparable performance to recent state-of-the-art CNN-based models on nominal relation classification (SemEval-2010 Task 8).  

Our evaluation and ablation led to three key findings.
First, the use of both word sequence and dependency tree structures is effective. 
Second, training with the shared parameters improves relation extraction accuracy, 
especially when employed with entity pretraining, scheduled sampling, and label embeddings. 
Finally, the shortest path, which has been widely used in relation classification, is also appropriate for representing tree structures in neural LSTM models.
 
\section*{Acknowledgments}
We thank Qi Li, Kevin Gimpel, and the anonymous reviewers for dataset details and helpful discussions.

\bibliography{acl2016}
\bibliographystyle{acl2016}

\appendix

\section{Supplemental Material}
\subsection{Data and Task Settings}

{\bf ACE05} defines 7 coarse-grained entity types: Facility ({\it FAC}), Geo-Political Entities ({\it GPE}), Location ({\it LOC}), Organization ({\it ORG}), Person ({\it PER}), Vehicle ({\it VEH}) and Weapon ({\it WEA}), and 6 coarse-grained relation types between entities: Artifact ({\it ART}), Gen-Affiliation ({\it GEN-AFF}), Org-Affiliation ({\it ORG-AFF}), Part-Whole ({\it PART-WHOLE}), Person-Social ({\it PER-SOC}) and Physical ({\it PHYS}).
We removed the {\it cts}, {\it un} subsets, and used a 351/80/80 train/dev/test split. We removed duplicated entities and relations, and resolved nested entities. We used head spans for entities. We follow the settings by \cite{li-ji:2014:P14-1}, and we did not use the full mention boundary unlike \newcite{lu-roth:2015:EMNLP}. We use {\it entities} and {\it
relations} to refer to {\it entity mentions} and {\it relation mentions} in ACE for brevity.

{\bf ACE04} defines the same 7 coarse-grained entity types as ACE05~\cite{doddington:5:2004:lrec2004}, but defines 7 coarse-grained relation types: {\it PYS}, {\it PER-SOC}, Employment / Membership / Subsidiary ({\it EMP-ORG}), {\it ART}, {\it PER}/{\it ORG} affiliation ({\it Other-AFF}), {\it GPE} affiliation ({\it GPE-AFF}), and Discourse ({\it DISC}).
We follow the cross-validation setting of \newcite{chan-roth:2011:ACL-HLT2011} and \newcite{li-ji:2014:P14-1}. We removed {\it DISC} and did 5-fold CV on {\it bnews} and {\it nwire} subsets (348 documents). We use the same preprocessing and evaluation metrics of ACE05.

{\bf SemEval-2010 Task 8} defines 9 relation types between nominals ( 
{\it Cause-Effect}, {\it Instrument-Agency}, {\it Product-Producer}, 
{\it Content-Container}, {\it Entity-Origin}, {\it Entity-Destination}, 
{\it Component-Whole}, {\it Member-Collection} and {\it Message-Topic}), and a tenth type {\it Other} when two nouns have none of these relations~\cite{hendrickx-EtAl:2010:SemEval}. 
We treat this {\it Other} type as a negative relation type, and no direction is considered.
The dataset consists of 8,000 training and 2,717 test sentences, and each sentence is
annotated with a relation between two given nominals.
We randomly selected 800 sentences from the training set as our development set. We followed the official task setting, and report the official
macro-averaged F1-score (Macro-F1) on the 9 relation types.

\subsection{Hyper-parameter Settings}

Here we show the hyper-parameters and the range tried for the hyper-parameters in parentheses. 
Hyper-parameters include the initial learning rate (5e-3, 2e-3, 1e-3, 5e-4,
2e-4, 1e-4), the regularization parameter (1e-4, 1e-5, 1e-6, 1e-7), dropout
probabilities (0.0, 0.1, 0.2, 0.3, 0.4, 0.5), the size of gradient clipping (1,
5, 10, 50, 100), scheduled sampling parameter $k$ (1, 5, 10, 50, 100), the number of
epochs for training and entity pretraining ($\leq$ 100), and the embedding dimension of WordNet hypernym (5, 10, 15, 20, 25, 30). 

\end{document}